\documentclass{article}

\usepackage[preprint]{neurips_2020}

\usepackage[utf8]{inputenc} 
\usepackage[T1]{fontenc}    
\usepackage{hyperref}       
\usepackage{url}            
\usepackage{booktabs}       
\usepackage{amsfonts}       
\usepackage{nicefrac}       
\usepackage{microtype}      
\usepackage{amsmath}
\usepackage{amssymb}
\usepackage{graphicx}
\usepackage{subfigure}
\usepackage{algorithm}
\usepackage{algorithmic}
\usepackage{multirow}
\usepackage{diagbox}
\usepackage{color}

\title{Multi-Objective Neural Architecture Search Based on Diverse Structures and Adaptive Recommendation}

\author{
  Chunnan Wang\textsuperscript{1}, Hongzhi Wang\textsuperscript{1,2}, Guosheng Feng\textsuperscript{1}\thanks{Hongzhi Wang, Chunnan Wang and Guosheng Feng contributed to this work equally.}, Fei Geng\textsuperscript{1} \\
  \textsuperscript{1}Harbin Institute of Technology\\
  \textsuperscript{2}Peng Cheng Laboratory\\
  \texttt{\{WangChunnan, wangzh, gengfei\}@hit.edu.cn, 1170300529@stu.hit.edu.cn} \\
}
\begin{document}

\maketitle

\begin{abstract}
The search space of neural architecture search (NAS) for convolutional neural network (CNN) is huge. To reduce searching cost, most NAS algorithms use fixed outer network level structure, and search the repeatable cell structure only. Such kind of fixed architecture performs well when enough cells and channels are used. However, when the architecture becomes more lightweight, the performance decreases significantly. To obtain better lightweight architectures, more flexible and diversified neural architectures are in demand, and more efficient methods should be designed for larger search space. Motivated by this, we propose MoARR algorithm, which utilizes the existing research results and historical information to quickly find architectures that are both lightweight and accurate. We use the discovered high-performance cells to construct network architectures. This method increases the network architecture diversity while also reduces the search space of cell structure design. In addition, we designs a novel multi-objective method to effectively analyze the historical evaluation information, so as to efficiently search for the Pareto optimal architectures with high accuracy and small parameter number. Experimental results show that our MoARR can achieve a powerful and lightweight model (with 1.9\% error rate and 2.3M parameters) on CIFAR-10 in 6 GPU hours, which is better than the state-of-the-arts. The explored architecture is transferable to ImageNet and achieves 76.0\% top-1 accuracy with 4.9M parameters.
\end{abstract}

\section{Introduction}\label{section:1}

Designing successful hand-crafted convolutional neural networks (CNN) is a laborious task due to the heavy reliance on expert experience and large amount of trials. To reduce the labour of human experts, neural architecture search (NAS) approaches~\cite{DBLP:conf/cvpr/ZophVSL18/NasNet,DBLP:conf/nips/NaymanNRFJZ19/XNAS,DBLP:conf/cvpr/ChenMZXHMW19/RENAS,DBLP:conf/iclr/LiuSY19/DARTS} are proposed to automatically discover effective CNN architectures. The main idea of existing NAS approaches is to define a search space and design a search strategy to find CNN architectures with high performance, e.g., high validation accuracy. Since the search space of CNN is huge~\cite{DBLP:conf/cvpr/DongY19/GDAS}, most NAS algorithms choose to use fixed outer network level structure, as is shown in Figure~\ref{fig1}, and search the repeatable cell structure only, so as to reduce searching cost. This kind of fixed structures perform well when enough cells and channels are used. However, when the architecture becomes more lightweight (with less parameters), its accuracy decreases significantly~\cite{DBLP:conf/cvpr/ZophVSL18/NasNet,DBLP:conf/nips/NaymanNRFJZ19/XNAS,DBLP:conf/icml/TanL19/ArcICML19,DBLP:journals/corr/abs-1904-04123/ASAP}. For example, when reducing the initial channel number from 44 to 36, the number of parameters in ASAP reduces 1.2M, and the test accuracy of ASAP on CIFAR-10 decreases 0.25\%~\cite{DBLP:journals/corr/abs-1904-04123/ASAP}.

Obviously, such inflexible structures prevent us from getting CNN with less parameters and higher accuracy. To get better lightweight architectures, we need to consider more flexible outer level structures and more diversified cell structures. MNasNet~\cite{DBLP:conf/cvpr/TanCPVSHL19/MNasNet} also noticed this problem. MNasNet pointed out that the cell structure diversity is significant in the resource-constrained CNN models, and studied more flexible CNN architectures, where each block of cells is allowed to contain different

\begin{minipage}[b]{0.45\linewidth}
structures and can repeat for different times. It searched the optimal setting of cell structures and cell numbers of different blocks, and achieved good results. Such solution breaks the traditional inflexible structures, but also has a defect, i.e. the search cost is too high. The search space of one cell is large, let alone that of more cells combined with parameters related to the outer level structures. The huge search space brings MNasNets considserable
\end{minipage}
\hfill
\makeatletter\def\@captype{figure}\makeatother
\begin{minipage}[b]{0.54\linewidth}
\centering
\includegraphics[height=7.75\baselineskip]{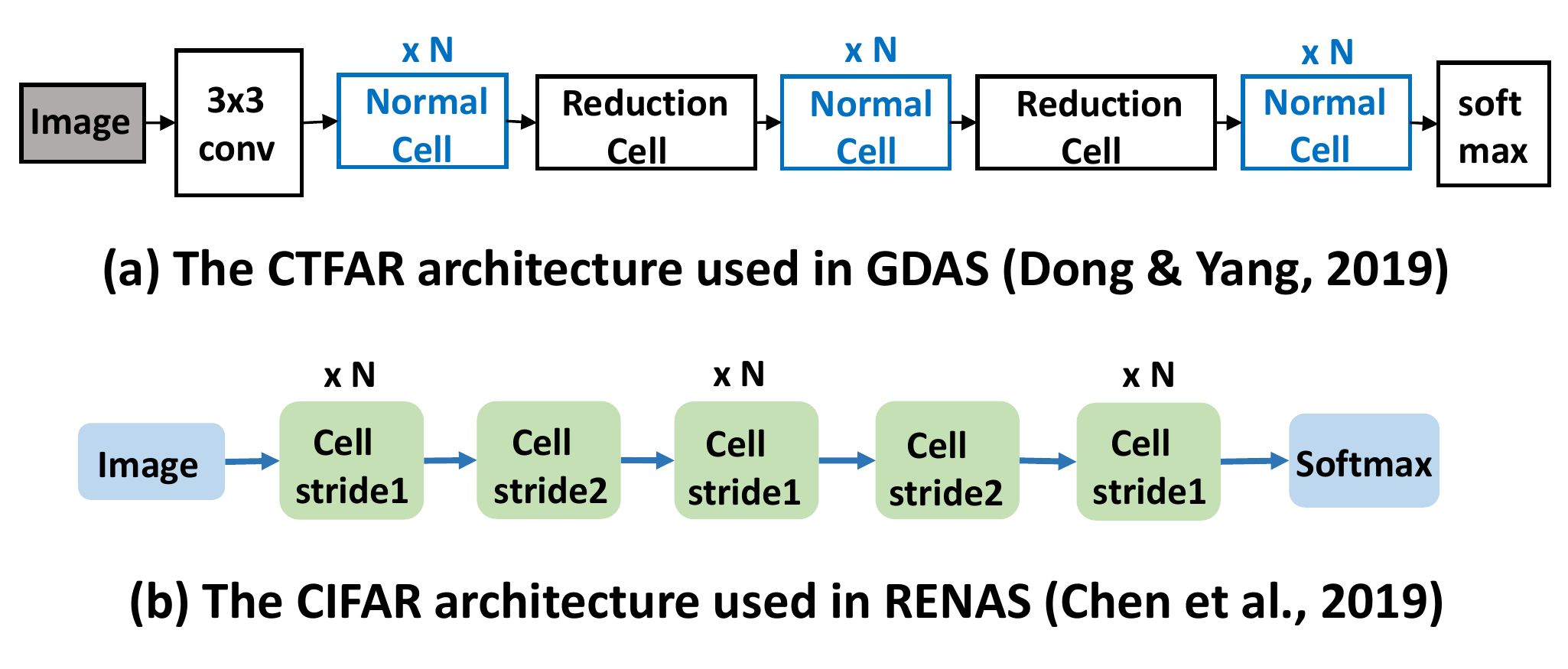}
\vspace{-0.35cm}
\caption{Examples of outer network level structures. Cell components are identical in these structures and are repeated to get deeper CNN.}
\vspace{-0.15cm}
\label{fig1}
\end{minipage}

search cost, i.e., MNasNet took hundreds of TPU days to accomplish the search process.

In order to control the size of search space thus to reduce search cost, and at the same time explore more flexible architectures for better models, i.e., Pareto optimal CNN models with high accuracy and small parameter number, in this paper, we put forward the idea of \underline{h}igh-performance \underline{c}ell \underline{s}tacking (HCS). That is, to utilize high-performance cells discovered by existing NAS algorithms to construct flexible architectures, as shown in Figure~\ref{fig2}, and search for the optimal cell stacking method to obtain better lightweight CNN. The introduction of existing high-performance cells, on the one hand, ensures the effectiveness of components, effectively reduces the search cost caused by cell design, and greatly reduces search space to avoid the search of invalid architecture, thus ensuring the search efficiency; on the other hand, increases the cell diversity as well as flexibility of CNN architectures. Our HCS-based search space could make full use of existing research results, and thus make it possible to explore more flexible CNN architectures efficiently, which is superior to existing search spaces.

In addition, in order to efficiently find Pareto optimal architectures that are lightweight and accurate in our newly designed search space, we design a multi-objective optimization (MOO) algorithm, called \underline{M}ulti-Objective \underline{O}ptimization based on \underline{A}daptive \underline{R}everse \underline{R}ecommendation (MoARR). The idea of MoARR is to avoid selecting worse architectures by effectively analyzing our historical evaluation information, thus reduce evaluation cost and accelerate the optimization. More specifically, MoARR utilizes historical information to study the potential relationship between the parameter quantity, accuracy and the architecture encoding, thus adaptively learn the \underline{r}everse \underline{r}ecommendation \underline{model} (RRModel) that is capable of selecting the most suitable architecture code according to the target performance. Then, MoARR recommends better architectures to be evaluated under the guidance of RRModel, i.e., inputting higher accuracy and smaller parameter number to RRModel for better architectures. With the increase of the evaluated architectures, RRModel becomes more reliable, and the architectures recommended by it approach to the Pareto Optimality. Using RRModel, MoARR can pertinently optimize architectures, and thus greatly reduce useless architecture evaluations.

Compared with the existing MOO approaches, our MoARR is more suitable for dealing with the MOO NAS problems, where architecture evaluations are expensive and time-consuming. More specifically, the existing approaches for seeking with Pareto-optimal front can be classified into two categories, approaches based on mathematical programming~\cite{DBLP:conf/nips/SenerK18/GredMO18} and those based on genetic algorithm~\cite{DBLP:journals/tec/ChengJOS16/MO16,DBLP:journals/tec/DebJ14/MO14,DBLP:journals/isci/QuS10/MO10,DBLP:journals/tec/DebAPM02/MO02}. The first class of methods are unable to cope with our black-box MOO NAS problem, where expression and gradient information of two optimization objective are unknown. The genetic methods can deal with the black-box problem, but could evaluate many useless architectures due to the uncertainties brought by many random operations and the neglect of valuable rules provided by historical evaluation information. They may require many samples and generations for good results, which is not suitable for dealing with expensive MOO NAS problems.


We compare MoARR with the classic NAS algorithms (Section~\ref{section:4}). Experimental results show that MoARR can find a powerful and lightweight model (with 1.9\% error rate and 2.3M parameters) on CIFAR-10 in 6 GPU hours, which outperforms the state-of-the-arts. The explored network architecture is transferable to ImageNet and 5 additional datasets, and achieves good results, e.g., 76.0\% top-1 accuracy with only 4.9M parameters on ImageNet. 

\section{Proposed Approach}\label{section:3}

\begin{figure}[t]
    \centering
    \includegraphics[width=0.96\columnwidth]{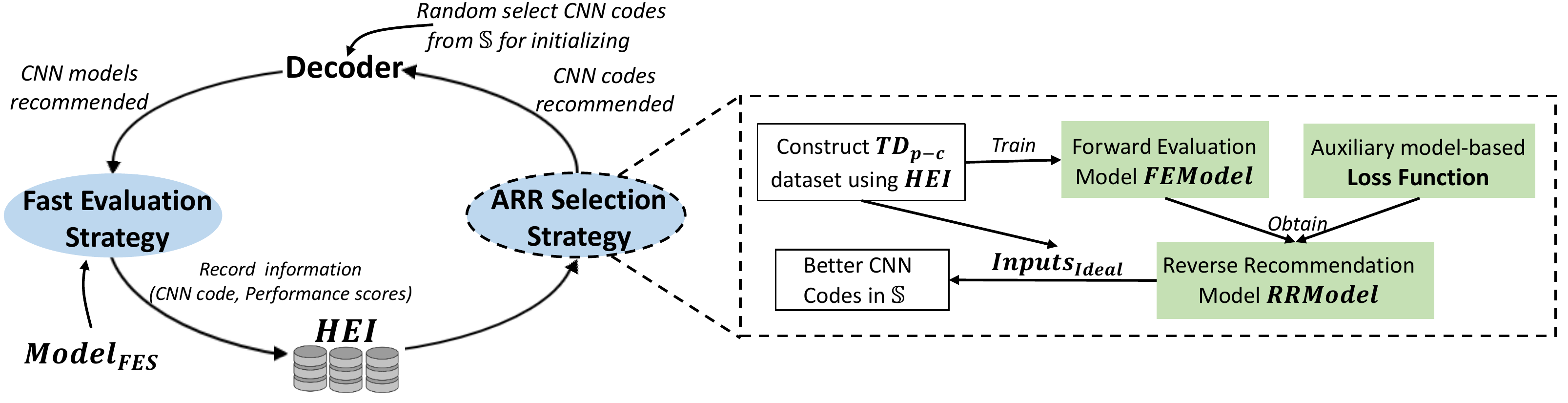}
    \vspace{-0.6cm}
    \caption{Overall framework of MoARR.}
\label{fig0}
\vspace{-0.5cm}
\end{figure}

To lay out our approach, we first give the specific definition of our research objective (Section~\ref{section:3.1}), and define a new search space of NAS (Section~\ref{section:3.2}). We then introduce MoARR that views NAS as a multi-objective optimization task, and makes full use of historical evaluation information to obtain high-performance light-weight CNN models (Section~\ref{section:3.3}). In order to accelerate the evaluation and reduce computational cost, we also design the acceleration strategy to use a small number of epochs and a few samples to quickly obtain accuracy scores (Section~\ref{section:3.4}). Figure~\ref{fig0} is our overall framework. 

\subsection{Target}\label{section:3.1}

In this paper, we aim to increase the flexibility and diversity of CNN architectures, so as to obtain lightweight architectures with higher accuracy. Formally, our search target is defined as follows:
\begin{equation}
\begin{split}
&\max \limits_{x\in \mathbb{S}}\,\, F(x)=[ACC(x), -PAR(x)]\\
&\ s.t. \quad Par(x) \leq P_{max}
\end{split}
\end{equation}
where $\mathbb{S}$ denotes all CNN architecture codes in our new search space, which is described in Section~\ref{section:3.2}, $ACC(x)$ denotes the accuracy score, $PAR(x)$ denotes the number of parameters, and $P_{max}$ is the upper limit of parameter amount. This is a multi-objective optimization task, 
and our goal is to obtain architectures that provide the best accuracy/parameter amount trade-off. 

\subsection{Search Space}\label{section:3.2}

\begin{minipage}[b]{0.45\linewidth}
Structural flexibility and cell diversity are two key points for the design of our search space. To achieve structural flexibility, we make the number of cells and channels in each stage to be adjustable. In this way, we can get architectures with diversified width and height. As for the cell diversity, we allow cells in different 
\end{minipage}
\hfill
\makeatletter\def\@captype{figure}\makeatother
\begin{minipage}[b]{0.54\linewidth}
\centering
\includegraphics[height=5.9\baselineskip]{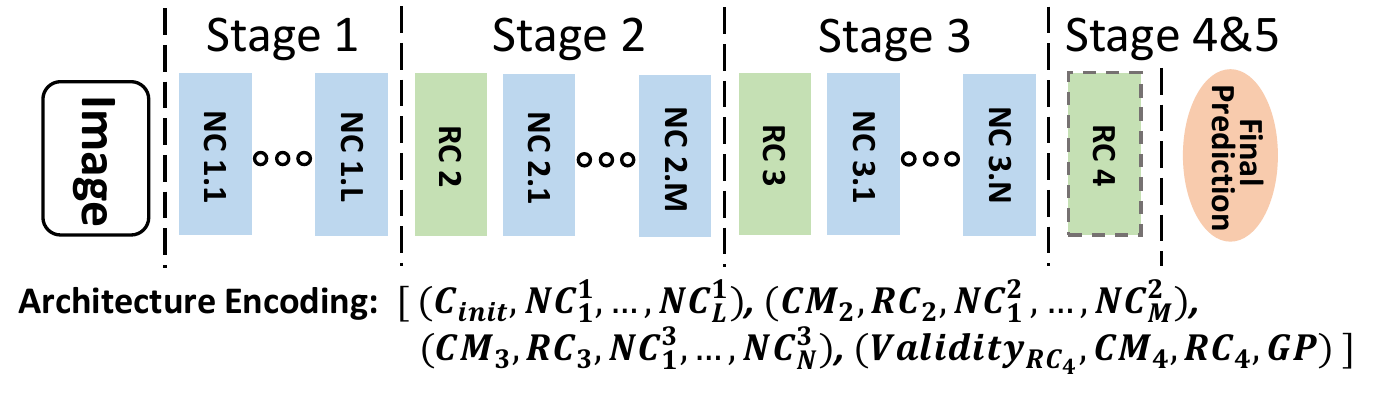}
\vspace{-0.7cm}
\caption{General CIFAR network architecture.}
\vspace{-0.15cm}
\label{fig2}
\end{minipage}

stages to have different structures, and take the existing high-performance cells structures discovered by previous NAS work as available options. Details are as follows.


We provide a general network architecture for our search space, as shown in Figure~\ref{fig2}. It consists of 5 stages. 
Stage 1 extracts common low-level features, stages 2$\textasciitilde$4 down-sample the spatial resolution of the input tensor with a stride of 2, and stage 5 produces the final prediction with a global pooling layer and a fully connected layer. Previous NAS approaches generally choose to use 2 reduction cells in CNN architectures, whereas some \cite{DBLP:conf/icml/PhamGZLD18/ENAS} uses 3. In pursuit of a more general search space, we use ${Validity}_{{RC}_{4}} \in \{True,\ False\}$ to decide whether to use the 3rd reduction cell in stage 4. For stages 1, 2, 3, each consists of $L$, $M$, $N$ normal cells, where $L$, $M$, $N$ are integers, i.e. $L, M, N \in \{4,5,6,7\}$. Different settings of L/M/N/$Validity_{RC_{4}}$ lead to different network depths. The width (number of output channels) of the cells in Stage 1 is denoted as $C_{init}\in \{112,128,144\}$, and that of the cells in Stage $s$ ($s=$2,3,4) is denoted by $C_{s}=C_{init}\times CM_{2}\times \ldots \times CM_{s}$, where $CM_{i}\in \{1.5,2,2.5\}$ represents the growth ratio of width compared to the previous stage. The name of the i-th normal cell in stage s is denoted as $NC_i^s$\footnote{The normal cells used in the same stage have the same name.}, the name of reduction cell used in stage s is denoted as $RC_s$, and the type of global pooling used in Stage 5 is denoted as $GP$. The options of $NC_i^s$, $RC_s$ and $GP$ are shown in Table~\ref{table1}. Therefore, an architecture can be encoded as shown in Figure~\ref{fig2}. The set of all possible codes is denoted as $\mathbb{S}$ and is referred to as our search space.

\begin{table}[t]
\newcommand{\tabincell}[2]{\begin{tabular}{@{}#1@{}}#2\end{tabular}}
\centering
\resizebox{0.615\textwidth}{!}{
\begin{tabular}{|l|l|l|l|}
\hline
\textbf{Cell Source} & \textbf{\tabincell{c}{Normal Cell Symbol}} & \textbf{\tabincell{c}{Reduction Cell Symbol}} \\
\hline
DARTS (1st)~\cite{DBLP:conf/iclr/LiuSY19/DARTS} & Darts\_V1\_NC & DARTS\_V1\_RC \\
DARTS (2nd)~\cite{DBLP:conf/iclr/LiuSY19/DARTS} & Darts\_V2\_NC & DARTS\_V2\_RC \\
NasNet-A~\cite{DBLP:conf/cvpr/ZophVSL18/NasNet} & NasNet\_NC & NasNet\_RC \\
AmoebaNet-A~\cite{DBLP:conf/aaai/RealAHL19/AmoebaNet} & AmoebaNet\_NC & AmoebaNet\_RC \\
ENAS~\cite{DBLP:conf/icml/PhamGZLD18/ENAS} & ENAS\_NC & ENAS\_RC \\
RENAS~\cite{DBLP:conf/cvpr/ChenMZXHMW19/RENAS} & RENAS\_NC & RENAS\_RC \\
GDAS~\cite{DBLP:conf/cvpr/DongY19/GDAS} & GDAS\_V1\_NC & GDAS\_V1\_RC \\
GDAS (FRC)~\cite{DBLP:conf/cvpr/DongY19/GDAS} & GDAS\_V2\_NC & GDAS\_V2\_RC \\
ASAP~\cite{DBLP:journals/corr/abs-1904-04123/ASAP} & ASAP\_NC & ASAP\_RC \\
ShuffleNet~\cite{DBLP:conf/cvpr/ZhangZLS18/ShuffleNet} & ShuffleNet\_NC & ShuffleNet\_RC \\
\hline
\end{tabular}
}

\resizebox{0.615\textwidth}{!}{
\begin{tabular}{|l|l|}
\hline
\textbf{Global Polling Definition} & \textbf{\tabincell{c}{Global Polling Symbol}} \\
\hline
Global average polling & Avg\_GP \\
Global max polling & Max\_GP \\
\tabincell{c}{The average of Avg\_GP and Max\_GP} & AvgMax\_GP \\
\hline
\end{tabular}
}

\caption{Options of $NC_{i}^{s}$, $RC_{s}$ and $GP$ in network architectures. We extract 10 normal cell structures and 10 reduction cell structures from 10 high-performance CNN architectures discovered by previous work, as the components. And we consider 3 classic global polling operations in the final stage.}
\label{table1}
\vspace{-0.9cm}
\end{table}

\subsection{MoARR: \underline{M}ulti-Objective \underline{O}ptimization based on \underline{A}daptive \underline{R}everse \underline{R}ecommendation}\label{section:3.3}

Let $x,y\in \mathbb{S}$ denote two elements in set $\mathbb{S}$. If $ACC(x)$$<$$ACC(y)$ and $PAR(x)$$>$$PAR(y)$, we say that architecture $y$ Pareto dominates $x$ ($y$ is better than $x$), denoted as $x\prec y$. For elements in set $\hat{\mathbb{S}}\subseteq \mathbb{S}$ that are not Pareto dominated by other elements in $\hat{\mathbb{S}}$, we call them the Pareto boundary of $\hat{\mathbb{S}}$, denoted as $B(\hat{\mathbb{S}})=\{x\in \hat{\mathbb{S}}\ |\ \nexists y\in \hat{\mathbb{S}},x\prec y\}$. Then, the Pareto optimal solutions for our multi-objective NAS problem is denoted by $B(\mathbb{S})$.

In MoARR, our target is to quickly optimize the elements in Pareto boundary $B(\hat{\mathbb{S}})$, where $\hat{\mathbb{S}}\subseteq \mathbb{S}$ denotes the set of evaluated architectures, and finally obtain $B(\mathbb{S})$. More specifically, we aim at selecting the best possible architectures to evaluate for each iteration, avoiding selecting worse architectures as much as possible, thus accelerate optimization process and reduce evaluation cost. To achieve this goal, we put forward \underline{A}daptive \underline{R}everse \underline{R}ecommendation (ARR), an architecture selection strategy which is capable of utilizing historical evaluation information of $\hat{\mathbb{S}}$ for effective and targeted architecture recommendation, i.e., recommending the most suitable architecture code according to the performance demands. Such performance-oriented architecture selection strategy can greatly reduce useless architecture evaluations and improve the quality of the selected architectures by setting superior performance scores, which is coincident with the goal of MoARR. Besides, ARR avoids the defects of genetic MOO methods mentioned in Section~\ref{section:1}, which makes MoARR more suitable for dealing with expensive NAS problem. We further discuss ARR as follows.

\textbf{ARR.} The model, which maps the performance pair ($acc$,$par$) to a suitable architecture code $x\ $$\in$$\ \{\ argmin_{x\in \mathbb{S}}\ $$(|ACC(x)-acc|^{2}$$+$$|PAR(x)-par|^{2})\ \}$ that has the closest performance scores, is called the \underline{r}everse \underline{r}ecommendation \underline{model} (RRModel) in ARR. And the core idea of ARR is to make full use of historical evaluation information $HEI\ $=$\ \{\ $$<$$x,ACC(x),PAR(x)$$>$ $|$ $x$$\in$$\hat{\mathbb{S}}\ \}$ to adaptively build effective RRModel, and then utilize RRModel to select superior architectures $SA\subseteq \{x\in \mathbb{S}\setminus \hat{\mathbb{S}}\ |\ \nexists y\in \hat{\mathbb{S}}, x\prec y\}$ directly by setting better performance values.


We note that the construction of effective RRModel is the key point of ARR. A straightforward solution for this task is to utilize $HEI$ to construct a performance-to-code training data $TD_{p-c}=\{\ $$<$$(acc,par),x$$>$$\ |\ $$<$$x,acc,par$$>$$\in$$HEI\ \}$, where the performance scores are considered as input and corresponding codes as the target output. Then use $TD_{p-c}$ to train a Multi-Layer Perception (MLP) to obtain RRModel. Note that there may exists different codes with totally the same or very similar accuracy score and parameter amount in $TD_{p-c}$, and the contradictory outputs may mislead the loss function and thus makes RRModel less effective. To eliminate the influence of contradictory values, this solution would only preserve one code for each performance pair in $TD_{p-c}$. However, such operation would also result in two defects: (1) Information loss, the valuable information contained in the deleted records is underutilized; (2) Difficulty in selection, how to select the most suitable code to preserve thus achieve the best recommendation effect is unknown, and many trails should be done to achieve the best results, which is time-consuming.

\begin{figure}[t]
\begin{minipage}[b]{0.58\linewidth}
\centering
\includegraphics[height=14.5\baselineskip]{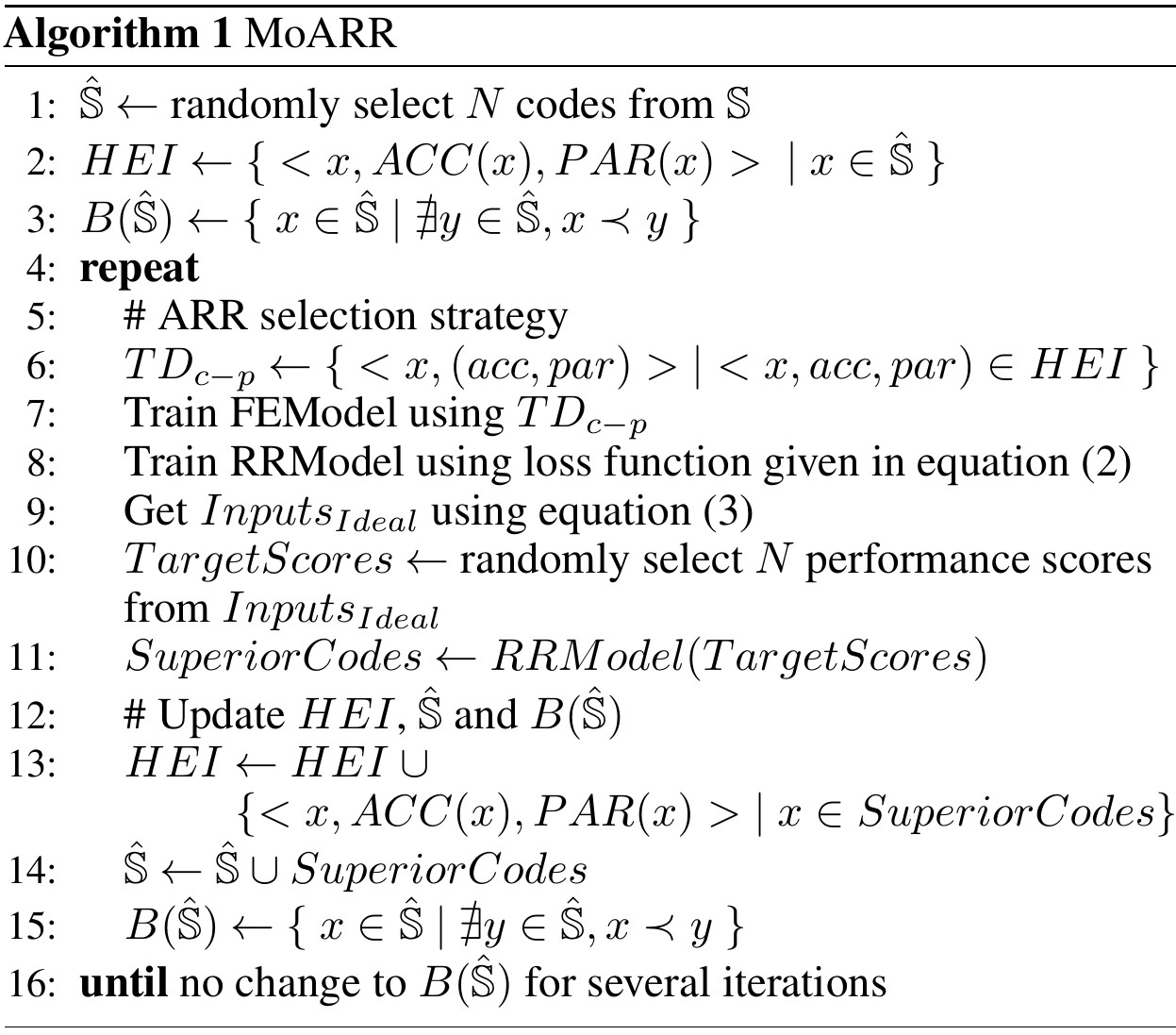}
\end{minipage}
\hfill
\makeatletter\def\@captype{figure}\makeatother
\begin{minipage}[b]{0.40\linewidth}
\centering
\includegraphics[height=13\baselineskip]{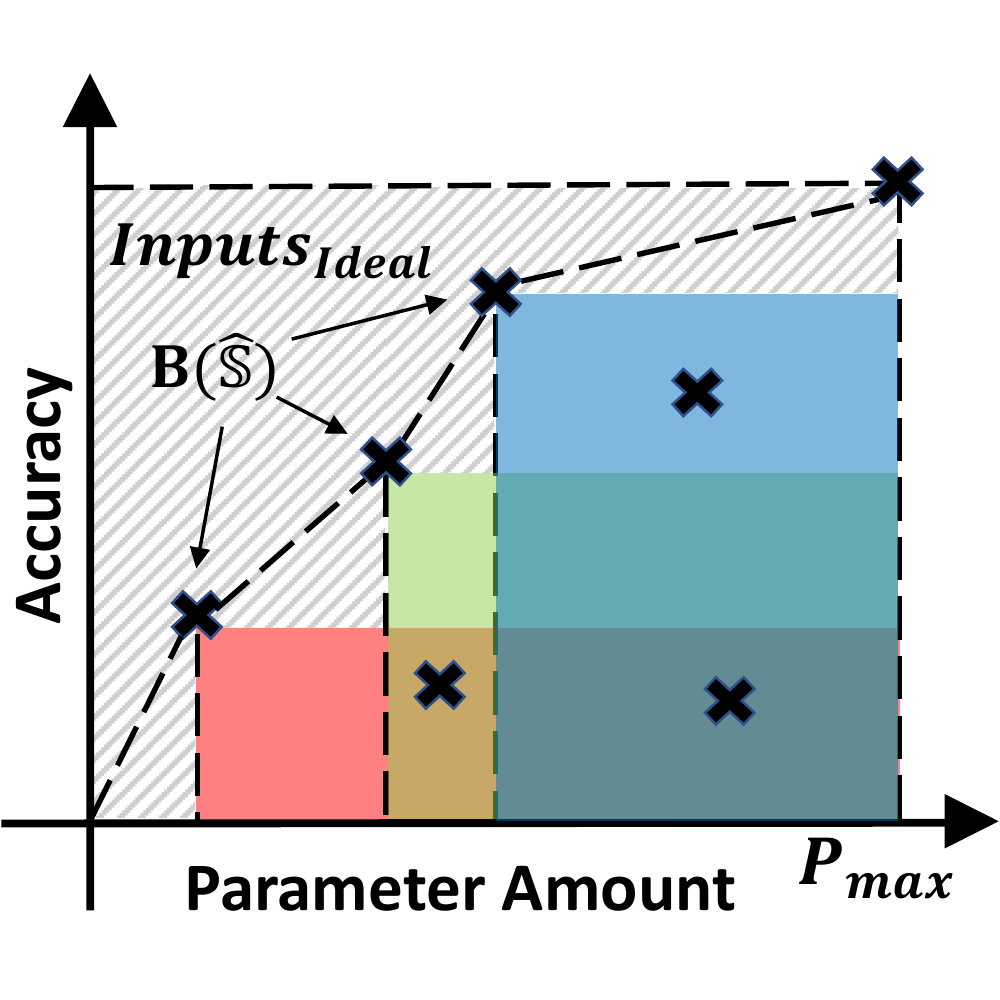}
\caption{An example of $Inputs_{Ideal}$.}
\label{fig3}
\end{minipage}
\vspace{-0.4cm}
\end{figure}

To avoid the above two defects, we propose an auxiliary model-based loss function for RRModel, which helps RRModel adaptively learn the most suitable output values by making full use of all historical information in $HEI$. Suppose \underline{f}orward \underline{e}valuation \underline{model} (FEModel) is capable of mapping the given architecture code to its performance pair, i.e., (accuracy score, parameter amount)\footnote{The mappings are opposite in FEModel and RRModel}. Then, the new loss function of RRModel is defined as follows:
\vspace{-0.25cm}
\begin{equation}
Loss(x,y')=\frac{1}{n} \times \sum_{i=1}^{n} ||x_{i}-FEModel(y_{i}')||^{2}
\label{equ1}
\end{equation}
where $x$$=$$\{x_1,...,x_n\}$ is a set of accuracy-parameter performance scores, and $y'=RRModel(x)$ denotes the suitable architecture codes recommended by RRModel. Equation~\ref{equ1} measures the differences between the target performance $x$ and the performance of the codes recommended by RRModel. It can help RRModel to automatically determine suitable outputs under the guidance of the auxiliary model FEModel. More specifically, we can input enough accuracy-parameter performance scores to RRModel without giving outputs, and RRModel can adjust its outputs adaptively according to the performance feedback provided by FEModel, and thus achieve reasonable recommendation. As for the FEModel, it is unknown since neural architecture evaluation is a black-box. However, we could utilize $HEI$ to construct code-to-performance training data $TD_{c-p}\ $=$\ \{\ $$<$$x,(acc,par)$$>$$\ |\ $$<x,acc,par$$>$$\in$$HEI\ \}$, and use $TD_{c-p}$ to train a MLP so as to approximate FEModel. Note that, different from $TD_{p-c}$, $TD_{c-p}$ does not exist contradictory problems. Therefore, with the usage of the new loss function, RRModel can be built automatically and effectively by making full use of all historical data $HEI$, and the two defectes are avoided.

The next step is to use the obtained RRModel to select superior architectures. Since the target is to optimize $B(\hat{\mathbb{S}})\ $=$\ \{\ x$$\in$$\hat{\mathbb{S}}$ $|$ $\nexists$ $y$$\in$$\hat{\mathbb{S}},$ $x$$\prec$$y\ \}$, we need to find the architecture codes that are not Pareto dominated by the evaluated codes $\hat{\mathbb{S}}$. Thus, we should input more competitive performance scores, i.e., performance scores with higher accuracy or lower parameter amount than the scores of $B(\hat{\mathbb{S}})$, to RRModel. We denote these performance scores as $Inputs_{Ideal}$, and its formula is given as follows:
\vspace{-0.10cm}
\begin{equation}
\begin{split}
Inputs_{Ideal}=\{(acc,par)\ |\ & 0<acc<1, 0<par<P_{max}, \\
&\forall x\in B(\hat{\mathbb{S}})\ \ ACC(x)\leq acc\ or\ PAR(x)\geq par\}
\end{split}
\label{equ2}
\end{equation}
Figure~\ref{fig3} is an example of $Inputs_{Ideal}$. Suppose shaped points are performance scores of elements in $\hat{\mathbb{S}}$, then shaded area is $Inputs_{Ideal}$. After getting $Inputs_{Ideal}$, we sample some superior performance scores randomly from $Inputs_{Ideal}$ as the inputs of RRModel, and thus get superior architectures. 

\textbf{MoARR.} With the usage of ARR, we develop MoARR algorithm, which deals with our multi-objective NAS problem effectively. Algorithm 1 is the pseudo code of MoARR. 

\subsection{Fast Evaluation Strategy}\label{section:3.4}

In MoARR, CNN code evaluation is very time-consuming due to the huge training dataset and large number of training epochs. 
In order to reduce the evaluation cost and thus speed up MoARR, we propose the \underline{f}ast \underline{e}valuation \underline{s}trategy (FES) to quickly estimate the final validation accuracy of CNN architectures in $\mathbb{S}$ using only a few training epochs and less training dataset.

\textbf{FES.} The core idea of FES is to use the following three types of characteristic attributes of an architecture $x\in \mathbb{S}$ to predict its $ACC_{Final}(x)$, which is the validation accuracy of $x$ obtained after $x$ is fully trained using all training dataset:
\begin{itemize}
\item Model complexity, including FLOPs and parameter amount of $x$;
\item Structural attributes, including Density, layer number and reduction cell number of $x$, where Density($x$) is the edge number divided by the dot number in DAG of $x$;
\item Quick evaluation scores, including top-1 accuracy, top-5 accuracy and loss value obtained by $x$ after training for 12 epochs using 1\% training dataset.
\end{itemize}
This attribute-based prediction method comes from the Easy Stop Strategy (ESS)~\cite{DBLP:conf/cvpr/ZhongYWSL18/BlockQNN}, which is successfully applied to CNN architectures, that are stacked with many copies of a discovered cell, to reduce the evaluation cost. In FES, we make some adjustments to ESS, making this method suitable for our more complex CNN architectures that are stacked with diversified cell structures. Our adjustments fall into two categories. Their reasons are as follows: (1) Replacing cell attributes to architecture ones. ESS uses the complexity and structural attributes of the used cell to stand for that of the whole architecture, and FES can only use that of the whole architecture to achieve the same description. (2) Involving more attributes. Our architectures are more flexible, and we use less training dataset for fast evaluation, thus we need more complexity features and structural attributes to distinguish different architectures, and need more performance features to substitute for $ACC_{Early}$.

To make this prediction method work for the architectures in our search space $\mathbb{S}$, in FES we sample some architectures from $\mathbb{S}$ to study the relationship between these attributes and $ACC_{Final}$. And finally we build a MLP regression model, denoted by $Model_{FES}$, to predict $ACC_{Final}(x)$ according to the above 8 attribute values of $x\in \mathbb{S}$. In MoARR, we utilize the obtained $Model_{FES}$ to efficiently estimate the $ACC_{Final}$ of candidate architectures recommended by ARR selection strategy. With the usage of $Model_{FES}$, we only cost about 20 seconds to estimate the final accuracy of an architecture $x\in \mathbb{S}$, which greatly reduces our evaluation cost and speeds up our algorithm.

\section{Experiments and Results}\label{section:4}

In this section, we test MoARR on common image classification benchmarks, and show its effectiveness compared to other state-of-the-art models (Section~\ref{section:4.1} to Section~\ref{section:4.3}). We use CIFAR-10~\cite{DBLP:/Cifar10} dataset for the main search and evaluation phase, and do transferability experiments on the well-known benchmarks using the architecture found on CIFAR-10. In addition, we conduct an ablation study which asserts the role of MoARR in discovering novel architectures (Section~\ref{section:4.4}).

\subsection{Details of Architecture Search on CIFAR-10}\label{section:4.1}

Using MoARR, we search on CIFAR-10~\cite{DBLP:/Cifar10} for better lightweight CNN architectures. Considering that CNN architectures discovered by existing NAS work are generally have more than 2.5M parameters~\cite{DBLP:conf/cvpr/DongY19/GDAS}, we set the upper limit of parameter amount $P_{max}$ in equation~\ref{equ1} to 2.5M. During the search phase, we select 50 architectures to evaluate for each iteration, and train each selected network for a fixed 12 epochs on CIFAR-10 using the FES as described in Section~\ref{section:3.4}. Following \cite{DBLP:conf/cvpr/ZhongYWSL18/BlockQNN}, we set the batch size to 256 and use Adam optimizer with $\beta_{1}$=0.9, $\beta_{2}$=0.999, $\varepsilon$=$10^{-8}$. The initial learning rate is set to 0.001 and is reduced with a factor of 0.2 every 2 epochs. Our MoARR takes about six hours to accomplish the search phase on a single NVIDIA Tesla V100 GPU.

After the search phase, we extract four excellent lightweight architectures from the Pareto boundary $B(\hat{\mathbb{S}})$ obtained by MoARR: (1) Two architectures with the highest accuracy score in $\{\ y$$\in$$B(\hat{\mathbb{S}})$ $|$ $2M\leq$$PAR(y)\leq$$2.5M\ \}$, which are represented by MoARR-Small1 and MoARR-Small2; (2) Two architectures with the highest accuracy score in $\{\ y$$\in$$B(\hat{\mathbb{S}})$ $|$ $PAR(y)$$<$$2M\ \}$, which are represented by MoARR-Tiny1 and MoARR-Tiny2. We then use these architectures (whose encodings are shown in the supplementary material) to test the effectiveness of MoARR.

\subsection{CIFAR-10 Evaluation Results}\label{section:4.2}

We train the MoARR-Small and MoARR-Tiny networks for 600 epochs using a batch size of 96 and SGD optimizer with nesterov-momentum and a weight decay of $3\times 10^{-4}$. We start the learning rate of 0.025 and reduce it to 0 with the cosine learning rate scheduler. For regularization we use cutout~\cite{DBLP:journals/corr/abs-1708-04552/CutOut}, scheduled drop-path~\cite{DBLP:conf/iclr/LarssonMS17/DropPath}, auxiliary towers~\cite{DBLP:conf/cvpr/SzegedyLJSRAEVR15/Tower} and randomly cropping. All the training parameters are the same as DARTS~\cite{DBLP:conf/iclr/LiuSY19/DARTS}. Table~\ref{table2} shows the performance of MoARR architectures compared to other state-of-the-art NAS approaches. From the experimental results, our MoARR-Small1 network

\begin{minipage}[b]{0.46\linewidth}
outperforms previous NAS methods by a large margin, with an error rate of 2.61\% and only 2.3M parameters. Moreover, MoARR-Small1 could reach 1.9\% error rate under the settings of ASAP~\cite{DBLP:journals/corr/abs-1904-04123/ASAP}, where 1500 epochs and more regularization methods are applied. Our smallest network variant, MoARR-Tiny2, outperforms most previous models on CIFAR-10, while having much less parameters, i.e., it contains 33.3\% to 94.4\% fewer parameters than previous models. With the consideration of more flexible and diversified structures, we discover CNN models with less parameters and higher accuracy using existing cell structures, which demonstrates the significance of cell diversity and structural flexibility. In addition, our MoARR is the second fastest among the
\end{minipage}
\hfill
\makeatletter\def\@captype{table}\makeatother
\begin{minipage}[b]{0.52\linewidth}
\newcommand{\tabincell}[2]{\begin{tabular}{@{}#1@{}}#2\end{tabular}}
\centering
\resizebox{\textwidth}{!}{
\begin{tabular}{|l|c|c|c|c|}
\hline
\textbf{\tabincell{l}{CIFAR-10\\ Architecture}} & \textbf{Venue} & \textbf{\tabincell{l}{Test\\ Error}} & \textbf{Params} & \textbf{\tabincell{l}{Search Cost\\ (GPU days)}} \\
\hline
PNAS~\cite{DBLP:conf/eccv/LiuZNSHLFYHM18/PNAS} & ECCV18 & 3.41 & 3.2M & 150 \\
AmoebaNet-A~\cite{DBLP:conf/aaai/RealAHL19/AmoebaNet} & AAAI19 & 3.12 & 3.1M & 3150 \\
DARTS (1st)~\cite{DBLP:conf/iclr/LiuSY19/DARTS} & ICLR19 & 3.00 & 3.3M & 1.5 \\
CARS-A~\cite{DBLP:journals/corr/abs-1909-04977/CARS} & CVPR20 & 3.00 & 2.4M & 0.4 \\
NAONet~\cite{DBLP:conf/nips/LuoTQCL18/NAONet} & NeurIPS18 & 2.98 & 28.6M & 200 \\
GDAS~\cite{DBLP:conf/cvpr/DongY19/GDAS} & CVPR19 & 2.93 & 3.4M & 0.21 \\
ENAS~\cite{DBLP:conf/icml/PhamGZLD18/ENAS} & ICML18 & 2.89 & 4.6M & 0.5 \\
RENAS~\cite{DBLP:conf/cvpr/ChenMZXHMW19/RENAS} & CVPR19 & 2.88 & 3.5M & 6 \\
CARS-B~\cite{DBLP:journals/corr/abs-1909-04977/CARS} & CVPR20 & 2.87 & 2.7M & 0.4 \\
GDAS (FRC)~\cite{DBLP:conf/cvpr/DongY19/GDAS} & CVPR19 & 2.82 & 2.5M & 0.17 \\
DARTS (2nd)~\cite{DBLP:conf/iclr/LiuSY19/DARTS} & ICLR19 & 2.76 & 3.3M & 4 \\
DATA~\cite{DBLP:conf/nips/ChangZGMXP19/DATA} & NeurIPS19 & 2.70 & 3.2M & 1 \\
NasNet-A~\cite{DBLP:conf/cvpr/ZophVSL18/NasNet} & CVPR18 & 2.65 & 3.3M & 1800 \\
CARS-I~\cite{DBLP:journals/corr/abs-1909-04977/CARS} & CVPR20 & 2.62 & 3.6M & 0.4 \\
ASAP~\cite{DBLP:journals/corr/abs-1904-04123/ASAP} & ArXiv19 & 1.99 & 2.5M & 0.5 \\
\hline
MoARR-Small1 & - & 2.61 & 2.3M & 0.27 \\
MoARR-Small2 & - & 2.69 & 2.2M & 0.27 \\
MoARR-Tiny1 & - & 2.74 & 1.9M & 0.27 \\
MoARR-Tiny2 & - & 2.76 & 1.6M & 0.27 \\
\hline
\end{tabular}
}
\vspace{-0.3cm}
\caption{Test error (\%) of MoARR compared to state-of-the-art methods on CIFAR-10.}
\label{table2}
\end{minipage}

NAS methods listed in Table~\ref{table2}, next to GDAS~\cite{DBLP:conf/cvpr/DongY19/GDAS}.

\subsection{Transferability Evaluation}\label{section:4.3}

Using the architecture found by MoARR searched on CIFAR-10, we preform transferability tests on 6 popular classification benchmarks.

\begin{minipage}[b]{0.46\linewidth}
\textbf{ImageNet Results.} Our ImageNet network is composed of two initial stem cells for downscaling and a new variant of MoARR-Small1 architecture for feature extraction and image classification. Following previous work, in the new variant of MoARR-Small1, we set $C_{init}$ to 184 and reduce 1 normal cell in each stage (stage 1 to 3), so the total number of network FLOPs is below 600M. Figure 3 from supplementary material shows our ImageNet architecture. We train the network for 250
\end{minipage}
\hfill
\makeatletter\def\@captype{table}\makeatother
\begin{minipage}[b]{0.52\linewidth}
\newcommand{\tabincell}[2]{\begin{tabular}{@{}#1@{}}#2\end{tabular}}
\centering
\resizebox{\textwidth}{!}{
\begin{tabular}{|l|c|c|c|c|c|}
\hline
\textbf{\tabincell{l}{ImageNet\\ Architecture}} & \textbf{\tabincell{l}{Top-1 Test\\ Error (\%)}} & \textbf{\tabincell{l}{Top-5 Test\\ Error (\%)}} & \textbf{Params} & \textbf{\tabincell{l}{Mult-\\Adds}} & \textbf{\tabincell{l}{Search Cost\\ (GPU days)}} \\
\hline
DARTS~\cite{DBLP:conf/iclr/LiuSY19/DARTS} & 26.9 & 9.0 & 4.9M & 595M & 1.5 \\
ShuffleNet (2x)~\cite{DBLP:conf/cvpr/ZhangZLS18/ShuffleNet} & 26.3 & - & 5.4M & 524M & - \\
GDAS~\cite{DBLP:conf/cvpr/DongY19/GDAS} & 26.0 & 8.5 & 5.3M & 581M & 0.21 \\
NasNet-A~\cite{DBLP:conf/cvpr/ZophVSL18/NasNet} & 26.0 & 8.4 & 5.3M & 564M & 1800 \\
PNAS~\cite{DBLP:conf/eccv/LiuZNSHLFYHM18/PNAS} & 25.8 & 8.1 & 5.1M & 588M & 150 \\
ENAS~\cite{DBLP:conf/icml/PhamGZLD18/ENAS} & 25.7 & 8.1 & 5.1M & 523M & 0.5 \\
DATA~\cite{DBLP:conf/nips/ChangZGMXP19/DATA} & 25.5 & 8.3 & 4.9M & 568M & 1 \\
AmoebaNet-A~\cite{DBLP:conf/aaai/RealAHL19/AmoebaNet} & 25.5 & 8.0 & 5.1M & 555M & 3150 \\
CARS~\cite{DBLP:journals/corr/abs-1909-04977/CARS} & 24.8 & 7.5 & 5.1M & 591M & 0.4 \\
ASAP~\cite{DBLP:journals/corr/abs-1904-04123/ASAP} & 24.4 & - & 5.1M & - & 0.2 \\
RENAS~\cite{DBLP:conf/cvpr/ChenMZXHMW19/RENAS} & 24.3 & 7.4 & 5.4M & 580M & 6 \\
\hline
MoARR-Small1 & \textbf{24.0} & \textbf{7.3} & 4.9M & 546M & 0.27 \\
\hline
\end{tabular}
}
\vspace{-0.3cm}
\caption{Transferability classification error on ImageNet dataset.}
\label{table3}
\end{minipage}
epochs with one cycle of the power cosine learning rate~\cite{DBLP:journals/corr/abs-1903-09900/CosineLR} and a nesterov-momentum optimizer. Results are shown in Table~\ref{table3}. We can observe from Table~\ref{table3} that MoARR's transferability results on ImageNet are highly competitive, outperforming all previous NAS models.

\textbf{Additional Results.} We further test MoARR transferability abilities on 5 smaller datasets: CIFAR-100~\cite{DBLP:/Cifar10}, Fashion-MNIST~\cite{DBLP:journals/corr/abs-1708-07747/FashionMnist}, SVHN~\cite{DBLP:SVHN}, Freiburg~\cite{DBLP:journals/corr/JundAEB16/Freiburg} and CINIC10~\cite{DBLP:journals/corr/abs-1810-03505/CINIC10}. We choose to use the MoARR-Small1 architecture, with similar training scheme as \cite{DBLP:journals/corr/abs-1904-04123/ASAP}. Table~\ref{table4} shows the performance of our model compared to other NAS methods. On Fashion-MNIST, MoARR-Small1 surpasses the next top architecture by 0.04\%, achieving the second highest reported score on Fashion-MNIST, second only to \cite{DBLP:conf/iclr/LiuSY19/DARTS}. On CIFAR-100, Freiburg and CINIC10, MoARR-Small1 surpasses all the other 6 architectures, achieving the lowest test errors.

\begin{table}[h]
\vspace{-0.3cm}
\newcommand{\tabincell}[2]{\begin{tabular}{@{}#1@{}}#2\end{tabular}}
\centering
\resizebox{0.86\textwidth}{!}{
\begin{tabular}{|l|c|c|c|c|c|c|c|}
\hline
\textbf{\tabincell{l}{Architecture}} & \textbf{\tabincell{c}{CIFAR-100\\ Test Error (\%)}} & \textbf{\tabincell{c}{FMNIST\\ Test Error (\%)}} & \textbf{\tabincell{c}{SVHN\\ Test Error (\%)}} & \textbf{\tabincell{c}{Freiburg\\ Test Error (\%)}} & \textbf{\tabincell{c}{CINIC10\\ Test Error (\%)}} & \textbf{\tabincell{c}{Params}} & \textbf{\tabincell{c}{Search Cost\\ (GPU days)}} \\
\hline
PNAS$^\dag$~\cite{DBLP:conf/eccv/LiuZNSHLFYHM18/PNAS} & 15.9 & 3.72 & 1.83 & 12.3 & 7.03 & 3.2M & 150 \\
AmoebaNet-A$^\dag$~\cite{DBLP:conf/aaai/RealAHL19/AmoebaNet} & 15.9 & 3.8 & 1.93 & 11.8 & 7.18 & 3.2M & 3150 \\
NasNet$^\dag$~\cite{DBLP:conf/cvpr/ZophVSL18/NasNet} & 15.8 & 3.71 & 1.96 & 13.4 & 6.93 & 3.3M & 1800 \\
NAONet~\cite{DBLP:conf/nips/LuoTQCL18/NAONet} & 15.7 & - & - & - & - & 10.6M & 200 \\
DARTS$^\dag$~\cite{DBLP:conf/iclr/LiuSY19/DARTS} & 15.7 & 3.68 & 1.95 & 10.8 & 6.88 & 3.4M & 4 \\
ASAP$^\dag$~\cite{DBLP:journals/corr/abs-1904-04123/ASAP} & 15.6 & 3.73 & 1.81 & 10.7 & 6.83 & 2.5M & 0.2 \\
\hline
MoARR-Small1 & \textbf{14.3} & 3.69 & \textbf{1.74} & \textbf{7.27} & \textbf{6.21} & 2.3M & 0.27 \\
\hline
\end{tabular}
}
\caption{Transferability classification error on 5 datasets.  Results marked with $^\dag$ are taken from \cite{DBLP:journals/corr/abs-1904-04123/ASAP}.}
\label{table4}
\vspace{-0.75cm}
\end{table}

\subsection{The Importance of MoARR}\label{section:4.4}

In this part, we analyze the importance of MoARR. We examine whether MoARR is actually capable of finding good CNN architectures, or whether it is the design of our new search space that leads to MoARR's strong empirical performance.

\textbf{Comparing to Guided Random Search.} We uniformly sample a CNN code from our search space, and build the CNN according to it, then we train this random model to convergence using the same settings mentioned in Section~\ref{section:4.2}. The random CNN model has 3.5M parameters and achieves the test error of 2.89\% on CIFAR-10 (using the setting of \cite{DBLP:journals/corr/abs-1904-04123/ASAP} yields 2.37\% error rate), which is far worse than MoARR-Small1’s 2.61\% and 1.90\%. It shows that our search space contains not only excellent lightweight architectures, but also many architectures with poor performance. Thus, efficient search strategy is necessary for our MOO NAS problem. The ARR selection strategy designed in MoARR can recommend good CNN codes with less parameters by analyzing the known performance information, which is effective.

\textbf{Comparing to Evolutionary Multi-Objective Search.} In addition to random search, we compare with the classic evolutionary multi-objective method RVEA*~\cite{DBLP:journals/tec/ChengJOS16/MO16}. We set the population size to 50, and use RVEA* instead to deal with our multi-objective NAS problem. Figure~\ref{fig4} reports the performance scores of architectures that are evaluated by RVEA* or MoARR in five generations. We
\begin{minipage}[b]{0.38\linewidth}
can observe that our MoARR evaluates much fewer useless architectures, and optimizes $B(\hat{\mathbb{S}})$ more quickly than RVEA*. Compared with the evolutionary method, our ARR selection strategy can recommend better architectures by utilizing potential relations learned from historical information. Our optimization process is more efficient and thus reduce the evaluation cost, which is more suitable to deal with the expensive multi-objective NAS problems, coinciding with the discussions in Section~\ref{section:1}.
\end{minipage}
\hfill
\makeatletter\def\@captype{figure}\makeatother
\begin{minipage}[b]{0.60\linewidth}
\centering
\includegraphics[height=11\baselineskip]{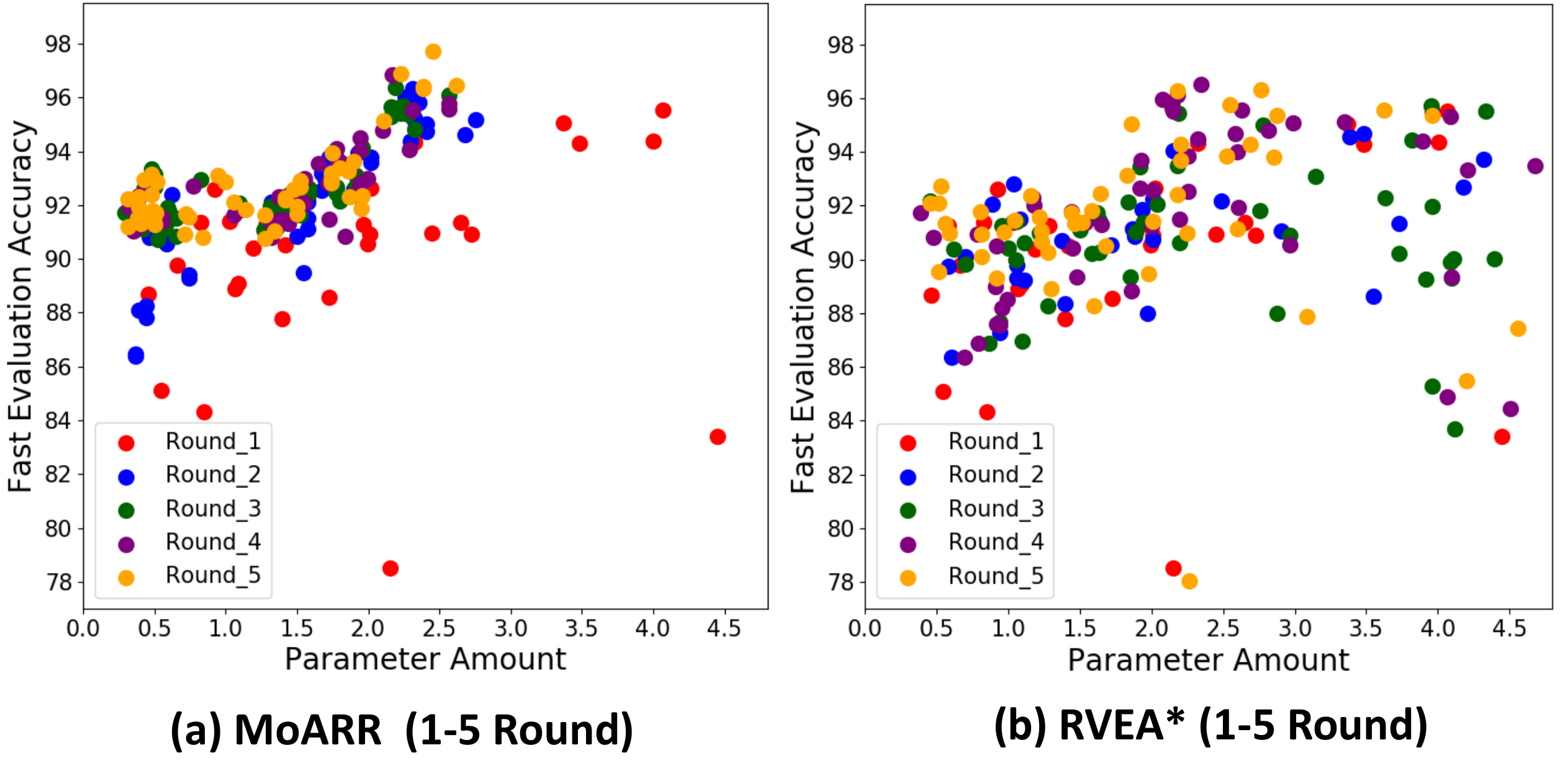}
\vspace{-0.6cm}
\caption{Performance of architectures evaluated by MoARR and RVEA*. Two algorithms use the same initial architectures, which are randomly-selected, in Round 1.}
\label{fig4}
\vspace{-0.1cm}
\end{minipage}


\section{Related Work}\label{section:2}

NAS is a popular and important research topic in deep learning. Many effective algorithms have been proposed to tackle this problem. Majority of them~\cite{DBLP:conf/cvpr/DongY19/GDAS,DBLP:conf/nips/NaymanNRFJZ19/XNAS,DBLP:conf/cvpr/ChenMZXHMW19/RENAS,DBLP:conf/iclr/LiuSY19/DARTS,DBLP:conf/icml/PhamGZLD18/ENAS,DBLP:conf/aaai/RealAHL19/AmoebaNet} adopt the idea of micro search, which centers on learning cell structures and designs a neural architecture by stacking many copies of the discovered cells, and the minority are macro search methods~\cite{DBLP:conf/icml/PhamGZLD18/ENAS,DBLP:conf/iclr/BakerGNR17/MacroICLR17,DBLP:conf/aaai/CaiCZYW18/MacroAAAI18,DBLP:conf/iclr/BrockLRW18/MacroICLR18}, which directly discover the entire neural networks. The former ones greatly reduce the computation cost but may miss some good architectures due to the inflexible network structure used by them, and the latter ones consider more flexible structures but are incapable of finding good architectures within short time due to the huge search space. In this paper, we propose to construct more flexible network structures utilizing good cell structures discovered by previous work, and thus efficiently search huger space for better CNN architectures. Our idea combines the merits of two methods and achieves better results.

More recently, with the increasing needs of deploying high-quality deep neural networks on real-world devices, multiple objectives are considered in NAS for real applications. Some works~\cite{DBLP:conf/cvpr/TanCPVSHL19/MNasNet,DBLP:conf/cvpr/WuDZWSWTVJK19/MOtoSO1,DBLP:conf/iclr/CaiZH19/MOtoSO2} tried to converted the multi-objective NAS tasks into the single-objective ones, and utilized the existing single-objective search methods, such as reinforcement learning~\cite{DBLP:conf/cvpr/ZophVSL18/NasNet,DBLP:conf/icml/PhamGZLD18/ENAS}, to deal with them. However, the weights in the single objective function are hard to determine, besides, the dimensional disunity of multiple objectives may result in poor robustness of single objective optimization. In this paper, we design MoARR to directly optimize multiple objectives, and thus avoid these problems.

\section{Conclusion and Future Works}\label{section:5}

In this paper, we propose MoARR for finding good lightweight CNN architectures. We construct more flexible and diversified network architectures using existing cell structures, and adaptively learn CNN recommendation model utilizing the performance feedback for efficiently optimizing architectures. Experimental results show that our MoARR can discover more powerful and lightweight CNN model compared with the state-of-the-arts, which demonstrates the importance of structural diversity and effectiveness of our optimization method. Our cell reusing idea and the multi-objective NAS optimization method are not only applicable to CNN but also other kinds of nerural networks. In the future works, we will further explore more various kind of network structures such as GNN and RNN, and improve the efficiency of MoARR.

\section*{Broader Impact}

To the best of our knowledge, we believe our work could benefit the society in areas that require image classification task. More specifically, our work could help to quickly generate high capacity models by utilizing current existing SOTA models when new problem come out and the image datasets are fresh. For example, to quickly generate robust chest CT image classification models after COVID-19 break out. Moreover, our lightweight model could deploy in light embedding devices, which make the technique more accessible for public. However, due to the possible system failure (e.g. misclassification), there may exist problems associated with public safety. For example, falsely classified products could enter the market which could cause serious problem, (e.g. unqualified medicine and agricultural products) and misclassified medical images may do harm to both the patients and the society. Such problem is hard to avoid due to the limited accuracy in current SOTA models as well as the unavoidable data quality problem, and we strongly hold that the model should be carefully adjusted and exhaustively tested (sometimes necessary manual assistance should be required) before putting into social practice.

\end{document}